\documentclass[sigconf]{aamas} 

\usepackage{balance} 

\usepackage[utf8]{inputenc} 
\usepackage[T1]{fontenc}    
\usepackage{hyperref}       
\usepackage{url}            
\usepackage{booktabs}       
\usepackage{amsfonts}       
\usepackage{nicefrac}       
\usepackage{microtype}      
\usepackage{xcolor}         
\usepackage{graphicx}
\usepackage{caption}
\usepackage{subcaption}
\usepackage{amsmath} 
\usepackage{esvect}
\usepackage{float}
\usepackage[linesnumbered, ruled, vlined]{algorithm2e}

\setcopyright{ifaamas}
\acmConference[AAMAS '23]{Proc.\@ of the 22nd International Conference
on Autonomous Agents and Multiagent Systems (AAMAS 2023)}{May 29 -- June 2, 2023}
{London, United Kingdom}{A.~Ricci, W.~Yeoh, N.~Agmon, B.~An (eds.)}
\copyrightyear{2023}
\acmYear{2023}
\acmDOI{}
\acmPrice{}
\acmISBN{}

\acmSubmissionID{136}

\title{The challenge of redundancy on multi-agent value factorisation}
\subtitle{Extended Abstract}

\author{Siddarth Singh}
\affiliation{
  \institution{Instadeep Ltd}
  \city{Johannesburg}
  \country{South Africa}}
\email{siddarthsingh2@gmail.com}

\author{Benjamin Rosman}
\affiliation{
  \institution{The University of the Witwatersrand}
  \city{Johannesburg}
  \country{South Africa}}
\email{benjamin.rosman1@wits.ac.za}

\begin{abstract}
In the field of cooperative multi-agent reinforcement learning (MARL), the standard paradigm is the use of centralised training and decentralised execution where a central critic conditions the policies of the cooperative agents based on a central state. It has been shown, that in cases with large numbers of redundant agents these methods become less effective. In a more general case, there is likely to be a larger number of agents in an environment than is required to solve the task. These redundant agents reduce performance by enlarging the dimensionality of both the state space and and increasing the size of the joint policy used to solve the environment.
  We propose leveraging layerwise relevance propagation (LRP) to instead separate the learning of the joint value function and generation of local reward signals and create a new MARL algorithm: relevance decomposition network (RDN). We find that although the performance of both baselines VDN and Qmix degrades with the number of redundant agents, RDN is unaffected.
\end{abstract}


\newcommand{\BibTeX}{\rm B\kern-.05em{\sc i\kern-.025em b}\kern-.08em\TeX}

\begin{document}


\pagestyle{fancy}
\fancyhead{}


\maketitle 


\section{Introduction}

For most multi-agent tasks in a practical setting, we would not know the precise number of agents required to optimally solve the problem. In general, there is likely to be a larger number of agents in an environment than is required to solve the task. As the number of independent agents increases so would the size of the state space that most problems require.

In complex environments constructing an accurate ground truth representation becomes difficult and in practical applications, the state representation data collected is often noisy or incomplete. Given the limitations of this space, it is not reliable to assume that the state space is reliable. When there is a large number of agents, many of them may be redundant for achieving an optimal policy. These redundant agents exacerbate the issue of the state space growth in the multi-agent setting. Therefore it is important to develop algorithms that can effectively separate agents that are essential to solving a task from the agents that are redundant to allow MARL algorithms to be deployed into more realistic and eventually real-world challenges.

We consider the idea of a collaborative task with a small margin of error like the traditional Piano Movers problem \cite{schwartz1983piano}. However, we consider the case where only $n$ of $m$ total agents in the environment are required to effectively complete the task. In the Piano Movers problem we therefore consider the case where an arbitrarily large number of agents are required to re-position the piano with each agent occupying a fixed amount of space in the environment along with the piano. Realistically with a large number of agents only a small subgroup $n$ of $m$ will be required to solve the task under the joint optimal policy. Ultimately during training, the policy will update to minimise the reliance on certain agents on the overall outcome as the joint optimal policy only require that they do not act in a manner that is destructive to the actions of the smaller group of required agents. In the Qatten \cite{qatten} paper the idea is put forward that both VDN and Qmix exhibit poor performance in environments with a high number of redundant agents as it is difficult to assign credit for task completion accurately when a disproportionately high credit is assigned to only a small number of the agents. 

In this paper, we propose a method to resolve this problem, relevance decomposition network (RDN) which makes use of layerwise relevance propagation (LRP) \cite{LRP} as an alternative to learned value decomposition only using local agent observations. By not using learned decomposition we can separate the learning of the joint value function from the training of the independent agents and make full use of the relationships between the local observations of the agents, their identities and their actions at each timestep to decompose the relationship between the global and local rewards. 

\section{Relevance Decomposition Network (RDN)}

We propose RDN to perform credit assignment between ad-hoc agents in the cooperative multi-agent setting under the assumption of a linear relationship between the individual local rewards and the shared global reward. However, unlike most central training decentralised execution (CTDE) methods which use learned decomposition as an end-to-end system, RDN separates the learning of the global reward function from the local Q-values of the agents \cite{MARL_OVERVIEW}. 

The main motivation for RDN is that in difficult settings, where agents are not all similarly responsible for the total reward, Qmix and VDN see decreased performance \cite{Qmix}\cite{VDN}. In these cases to achieve accurate reward assignment, some agents must be weighted much higher than others when performing value decomposition \cite{qatten}. In the case of many redundant agents, Qmix and VDN have difficulties decomposing the relationship between the high and low-value agents \cite{SMAC}. Due to this, the decomposed rewards have high variance during training which makes discovering stable policies difficult. LRP is more effective at determining this relationship, which stabilises the training process.

For RDN our independent agents are modelled as Deep Q Networks (DQNs) \cite{DQN}. These DQNs gather data where $Q^i(o_{t,i,.})$ is the $Q$ value of independent agent $i$ at timestep $t$, $h^i_t$ is the hidden state of agent $i$ at timestep $t$, $o_{t,i,.}$ is the local observation of agent $i$ at timestep $t$,$a_{t,i}$ is the action taken by agent $i$ at timestep $t$ and $\pi_i(Q^i(o_{t,i,.}),\epsilon (e))$ is the policy $\pi$ of agent $i$ dictated by a Q value function and an $\epsilon-greedy$ exploration strategy. We use the critic network to calculate the expected total Q value at each timestep parameterised by $ \theta^c$ and a target Q value using a target critic parameterised by $\Tilde{\theta^c}$ whose parameters are copied over from the critic at fixed intervals. The critic network is updated using loss $L(\theta^c )= E_{\vv{o},\vv{a},r,\vv{o}'}[(Q^{\theta^c}_{tot}(o_1,...,o_n,a_1,...,a_n)-y)^2]$ where $y = r + \gamma(Q^{{\theta'}^c}_{tot}(o^{'}_1,...,o^{'}_n,a^{'}_1,...,a^{'}_n))$ and $\theta^c$ is the critic's parameters and $\theta^{'c}$ is the target critic parameters, which are reset every $C$ training epochs. To train the agents, we calculate the Q value at the current timestep for each independent agent. This is done in the same manner as standard independent Q-learning where we calculate $Q(o,a_i)$ for each agent based on their local observations and actions. We calculate the total target Q-value for the current timestep $t$ using a separate critic network. This critic takes in the concatenation of local agent observations and actions to predict the target total Q-value of the global state at the current timestep then we use LRP to decompose the global reward calculated into local rewards for each agent. Essentially we assign relevance scores to all data points in the concatenated observation space and then separate the scores for each index based on the agent they were collected or generated from. The concatenated observation space is a concatenation of the local agent observations and the actions of each agent at each timestep. The relevance scores for each agent are then summed per agent and are used as the target Q-values to train the independent learners as if performing standard independent Q learning. Finally, the total expected Q value for each timestep is decomposed into independent target Q-values $\Tilde{\theta^i}$ and the DRQNs \cite{drqn} which act as the agent networks are trained using the loss $L(\theta^i) = E_{\vv{o},\vv{a},r,\vv{o}'}[(Q^{i,\theta^i}(o_i,a_i)-\Tilde{Q}^{i})^2]$ \cite{sandholm1995multiagent} and $\Tilde{Q}^{i} =  \sum_i R_{in}$ where $\Tilde{Q}^{i}$ is the $Q$ target for agent $i$ and $\sum_i R_{in}$ is the sum of all relevance values associated with agent $i$.

A characteristic of LRP is it maintains a conservative calculation between layers of the neural network \cite{LRP}. As such the relevance values from later layers are included completely in the calculation of the relevance values of the layers closer to the input. Essentially we can assume that the total sum of the relevance values is approximately the same as the output of the NN when LRP is used. Therefore we can equate $Q_{tot}$ to the sum of relevance scores as $Q_{tot} \approx  R_{in} $.

\section{Results}

The pre-existing map from \textit{the Starcraft Multi-Agent Challenge} (SMAC) we make use of is the \textit{bane\textunderscore vs\textunderscore bane} map. In this map, each side has 20 zerglings and 4 banelings. The most optimal policy for this environment has the zerglings move out of the way to not obstruct the banelings' movement. We make use of 3 additional variants of the original map. \textit{bane\textunderscore med} which only has 15 zerglings, \textit{bane\textunderscore small} with has 10 zerglings and \textit{ bane\textunderscore no\textunderscore z} with has no zerglings. We vary the number of redundant agents (the zerglings) to show the effect of redundancy on performance.

\begin{figure}[h]
    \centering
  \includegraphics[width=0.9\linewidth]{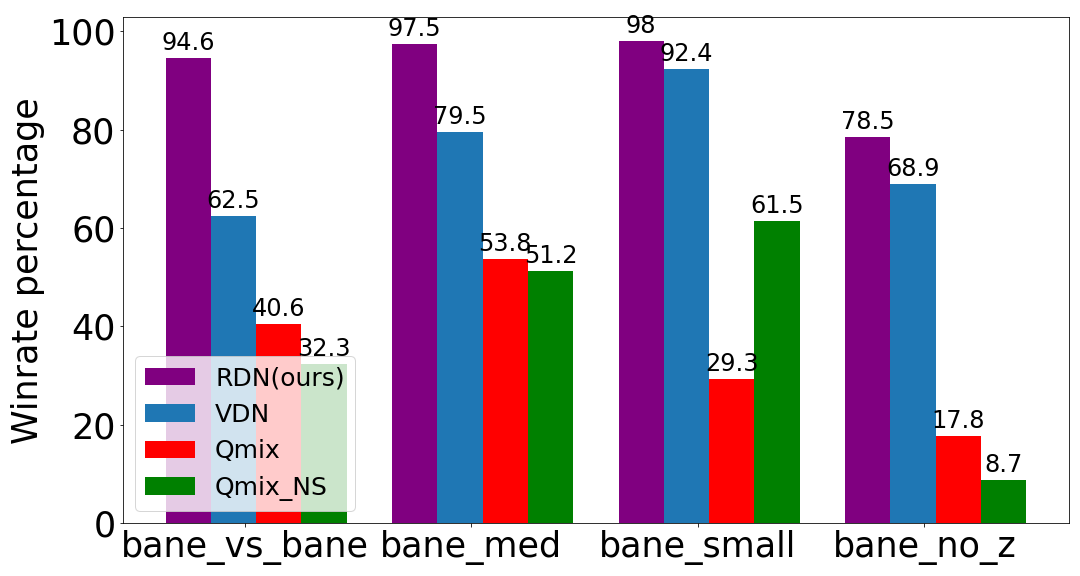}
  \caption{Percentage winrates from highest number of redundant agents (left) to least redundant agents (right) for all tested algorithms } 
  \label{fig:percentages}
\end{figure}

From figure \ref{fig:percentages} we can see that across all maps RDN outperforms all baselines although the performance of VDN improves as the number of redundant agents is reduced. Interestingly QMIX is outperformed by QMIX\textunderscore NS which does not use the global state in \textit{bane\textunderscore med} and in \textit{ bane \textunderscore no\textunderscore z} indicating that in cases where there are an intermediate number of redundant agents the central state already begins to become uninformative making accurate multi-agent credit assignment difficult.

In the case where all redundant agents are removed VDN performs similarly to RDN however, we also find that when no redundant agents are present at all in the \textit{ bane \textunderscore no\textunderscore z} map performance decreased across all algorithms when compared to \textit{bane\textunderscore small}. We suspect that having a small number of extra agents may aid in exploration early in the training regime.

\section{Conclusion}

We show how increasing numbers of redundant agents makes reaching stable convergence to an optimal joint policy difficult for both monotonic factorisation methods like QMIX and linear factorisation methods like VDN where only $n$ of $m$ total agents are required for an environment to be solved. We then propose RDN as a method that uses LRP to perform more optimal credit assignments in environments with high numbers of redundant agents using only local agent observation. RDN can reach near-optimal convergence on all environments used with similar overall winrates without the use of the ground truth state information. With VDN and QMIX we see a gradual decay in performance and, an increase in variance as the number of redundant agents increases.

\bibliographystyle{IEEEtran}
\bibliography{main}
\end{document}